\documentclass[runningheads]{llncs}
\usepackage{graphicx}

\usepackage{tikz}
\usepackage{comment}
\usepackage{enumitem}
\usepackage{xspace}
\usepackage{amsmath,amssymb} %
\usepackage{color}
\usepackage{float}
\usepackage[accsupp]{axessibility}  %
\usepackage{pifont}
\usepackage{tikz}
\usepackage{pgfplots}
\usepackage{orcidlink}
\usepackage{algorithm}
\usepackage[noend]{algpseudocode}
\usepackage{makecell}

\makeatletter
\newcommand{\@chapapp}{\relax}%
\newcommand{\printfnsymbol}[1]{%
  \textsuperscript{\@fnsymbol{#1}}%
}

\makeatother

\newcommand*{\rev}{\textcolor{black}}

\newcommand{\bx}{\boldsymbol{x}}

\newcommand{\bu}{\boldsymbol{u}}

\newcommand{\sY}{\mathcal{Y}}

\newcommand{\bz}{\boldsymbol{z}}

\newcommand{\bv}{\boldsymbol{v}}

\newcommand{\sS}{\mathcal{S}}
\newcommand{\sT}{\mathcal{T}}

\newcommand{\sI}{\mathcal{I}}

\DeclareMathOperator*{\argmin}{arg\,min}

\newcommand{\field}[1]{\mathbb{#1}}

\newcommand{\R}{\field{R}}

\makeatletter

\usepackage{accents}

\usepackage{xcolor}
\usepackage{multirow}

\makeatother

\newcommand{\OUR}{ReSeND\xspace}

\begin{document}
\pagestyle{headings}
\mainmatter
\def\ECCVSubNumber{5215}  %

\title{Semantic Novelty Detection \\via Relational Reasoning}

\titlerunning{Semantic Novelty Detection via Relational Reasoning}

\author{Francesco Cappio Borlino\thanks{equal contributions}\inst{,1,2}\orcidlink{0000-0002-8507-0213} \and
Silvia Bucci$^{\star}$\inst{,1}\orcidlink{0000-0001-6318-7288} \and
Tatiana Tommasi\inst{1,2}\orcidlink{0000-0001-8229-7159}}

\authorrunning{F. Cappio Borlino et al.}
\institute{Politecnico di Torino, Corso Duca degli Abruzzi 24, 10129 Torino, Italy \and
Italian Institute of Technology, Italy
\email{\{francesco.cappio,silvia.bucci,tatiana.tommasi\}@polito.it}}
\maketitle
\begin{abstract}
Semantic novelty detection aims at discovering unknown categories in the test data. This task is particularly relevant in safety-critical applications, such as autonomous driving or healthcare, where it is crucial to recognize unknown objects at deployment time and issue a warning to the user accordingly.
Despite the impressive advancements of deep learning research, existing models still need a finetuning stage on the known categories in order to recognize the unknown ones. This could be prohibitive when privacy rules limit data access, or in case of strict memory and computational constraints (e.g. edge computing). 
We claim that a tailored representation learning strategy may be the right solution for effective and efficient semantic novelty detection. Besides extensively testing state-of-the-art approaches for this task, we propose a novel representation learning paradigm based on relational reasoning. It focuses on learning how to measure semantic similarity rather than recognizing known categories.
Our experiments show that this knowledge is directly transferable to a wide range of scenarios, and it can be exploited as a plug-and-play module to convert closed-set recognition models into reliable open-set ones.

\keywords{Representation Learning, Novelty Detection, Open Set Learning, Domain Generalization, Relational Reasoning}
\end{abstract}

\section{Introduction}

In the last years, deep learning models have brought significant advances in several computer vision tasks. 
We can identify two main ingredients \rev{as} the basis of this widespread success.
The first one is the pre-training stage: the possibility to rely on a large set of freely available images %
allows to learn a representation that is generally helpful to initialize the models. The second component is the optimistic assumption that training and test distributions 
will perfectly match. 
Indeed, in real-world conditions,
it's much more common to encounter 
differences between the two, for instance\rev{,} due to a mismatch among their semantic category sets.
This condition is particularly 
dangerous in safety-critical applications like autonomous driving and healthcare, where %
previously unseen categories should be reliably detected as \textit{unknown}. 
Several studies have proposed to improve the learning procedure and make it aware of semantic novelties outside of the training distribution. Existing solutions consist in calibrating the softmax output of deep classifiers \cite{hendrycks17baseline,liang2017enhancing,liu2020energy}, or using generative approaches to synthesize outliers \cite{Neal_2018_ECCV,GenOpenMax,aaai_SensoyKCS20,xia2020synthesize,nalisnick2018do}. 
However, a relevant limitation of these %
techniques %
is that all of them require to be trained, or at least finetuned, on a reasonably large set of %
reference data in order to learn what is \textit{known}. %
In case of limited data access due to privacy concerns,  
or when dealing with memory and computational constraints (e.g. edge computing), these strategies could be inapplicable.

\begin{figure}[tb]
\centering
\includegraphics[width=0.98\textwidth]{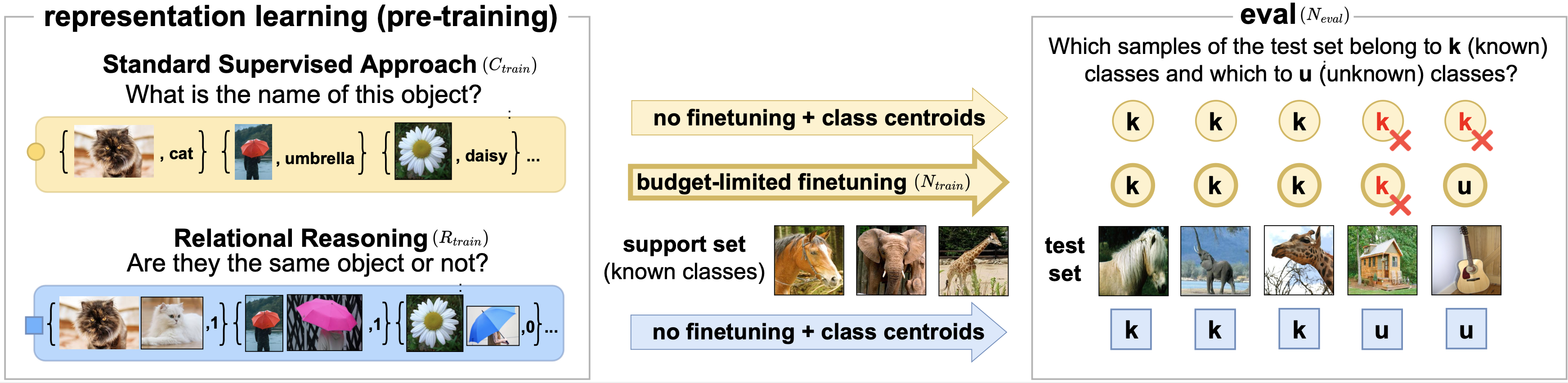}
    \caption{
    Comparison between standard supervised learning and relational reasoning representation learning. 
    The first aims at recognizing the known object classes, while the second learns a measure of semantic similarity among image pairs. %
    We claim and verify experimentally that relational reasoning is particularly suitable when the final goal is semantic novelty detection. Our pre-trained large-scale relational model can be transferred on semantic novelty detection tasks without the need for a finetuning phase on the known classes of the task at hand.
    }   \vspace{-3mm}
    \label{fig:teaser}
\end{figure}

In this work, we put the spotlight on the pre-training stage.
We claim that, rather than considering the usual cross-entropy based classification \cite{he2016deep}, or self-supervised contrastive learning \cite{simclr2020,MOCO_He_2020_CVPR}, we can exploit ImageNet1k to optimize a relational reasoning objective and obtain a more reliable embedding for novelty detection  (see Fig. \ref{fig:teaser}). 
Specifically, our target is a 
semantic similarity measure that indicates whether two samples belong to the same class or to different ones. %
Thus, we focus on learning a representation designed for semantic comparison which does not need further finetuning on the annotated data of the task at hand. It will be enough to compare each test sample with the reference class-prototypes to separate known and unknown categories. 
Besides being an efficient strategy, our method provides a plug-and-play solution to convert existing closed-set models to open-set ones by including a rejection option for unknown classes.

To summarize, \textbf{we focus on Semantic Novelty Detection (SeND) and propose \OUR, a representation learning approach based on Relational Reasoning that is ready to be used in real-world applications without the need for finetuning}.
In particular, our contributions are:
\begin{itemize}[leftmargin=*]
\item we conduct a thorough experimental analysis on the ability of several representation learning paradigms %
to deal with the SeND task, exploring their potentialities and limits;
\item we introduce \OUR and evaluate it
on several \textit{intra-} and \textit{cross-}domain scenarios, exploring settings with different ratios of unknown classes in the test data. 
An extensive benchmark with several competitors confirms the effectiveness and efficiency of our approach;

\item we show how \OUR can be used as a plug-and-play module on closed-set domain generalization approaches converting them into open-set domain generalization strategies that set the new state-of-the-art.

\end{itemize}

\section{Related Works}
Our work relates to three main research areas: representation learning, relational reasoning, and out-of-distribution detection. 

\medskip\noindent\textbf{Representation Learning} makes the difference between classic shallow and modern deep machine learning approaches. The former relies on handcrafted feature representation, while the latter automatically learns to represent the input data through a hierarchy of features during the training process. The literature on this topic is quite extensive \cite{BengioTPAMI2013,GoodBengBook}, ranging from the design of neural architectures \cite{HintonSalakhutdinov2006b,Kingma2014,NIPS2014gan} to the development of learning paradigms \cite{simclr2020,caron2020unsupervised,du2021curious}. 
The most common approach used to get effective representations from visual data is supervised learning, but recent works have been mainly dedicated to learning representations from unlabeled samples \cite{gidaris2018unsupervised,noroozi2016,Jenni_steering,yun2019cutmix,cutpaste,simclr2020,MOCO_He_2020_CVPR,barlow_ZbontarJMLD21,Chen_2021_CVPRsimsiam}. They showed how the obtained self-supervised embeddings are able to capture general knowledge \rev{of} data structure and can be leveraged by a large variety of downstream tasks \cite{kolesnikov2019revisiting,Newell_2020_CVPR,Ericsson2021HowTransfer}. Usually, this happens via a transfer learning procedure that requires finetuning on annotated training data of the final task. 

\medskip\noindent\textbf{Relational Reasoning} is a hallmark of human intelligence and it has been formalized by the machine learning community as learning a function to quantify the relationships between a set of objects. This paradigm has attracted particular attention for the combination of language and vision for scene description %
\cite{Johnson_clevr,Santoro_nips17,Raposo_iclr17}. 
Other applications are on %
reinforcement learning  \cite{Santoro_nips18,pan2021actor,zambaldi2018deep}, object detection \cite{Hu_2018_CVPR}, graph networks \cite{relationalbias_arxiv}, and few-shot learning \cite{sung2018_LearningToCompare,Zhang_2021_CVPR}. 

\noindent\emph{Relational reasoning and contrastive learning}. Recently, it has been shown that relational reasoning can effectively guide self-supervised representation learning \cite{patacchiola2020_SELFrelreason}, with better results than those of popular contrastive learning strategies \cite{simclr2020,hjelm2018learning}.
On the basis of these results, we can identify one important aspect that makes relational reasoning different from contrastive learning. \rev{The latter aims at learning a feature space for individual samples, with the similarity between two samples computed a posteriori using a distance metric; the goal of the former is to construct a representation for sample pairs: the position of a point in the final embedding directly represents the similarity between two samples.}

\medskip\noindent\textbf{Out-Of-Distribution detection (OOD)} studies how to identify whether a given test sample is drawn from the training distribution or not.
Both a variation in semantic content and in the visual domain may cause a deviation from the reference distribution. OOD is a wide framework that covers several sub-settings.

\smallskip\noindent\emph{OOD subsettings}. In \emph{anomaly detection} the training samples belong to a single semantic category and a test sample is considered anomalous both if it contains a novel class and in case it presents the same known class but with perceptual differences from the training (e.g. local defects, global style). 
When the training data cover more than one class, the setting is usually indicated as \emph{novelty detection}. %
As in anomaly detection, the cause of novelty can be either a semantic shift, or domain shift or both \cite{hendrycks2018deep,Yang_2021_ICCV}. We use the name \emph{semantic novelty detection} (SeND) to focus on the first case: models that spot unknown categories in the test while being agnostic to domain variations \cite{oza2020multiple}. 
\emph{Open-set recognition} extends novelty detection by considering not only a binary identification of known and unknown classes in the test, but also a reliable recognition \rev{of} the known classes. Usually, this setting is well controlled with training and test data sharing the same visual domain. In \emph{open-set domain generalization} the %
model should be also robust to the domain shift between train and test data \cite{shu2021_OPENDG}.

\begin{figure*}[tb]
\centering
\includegraphics[width=0.95\textwidth]{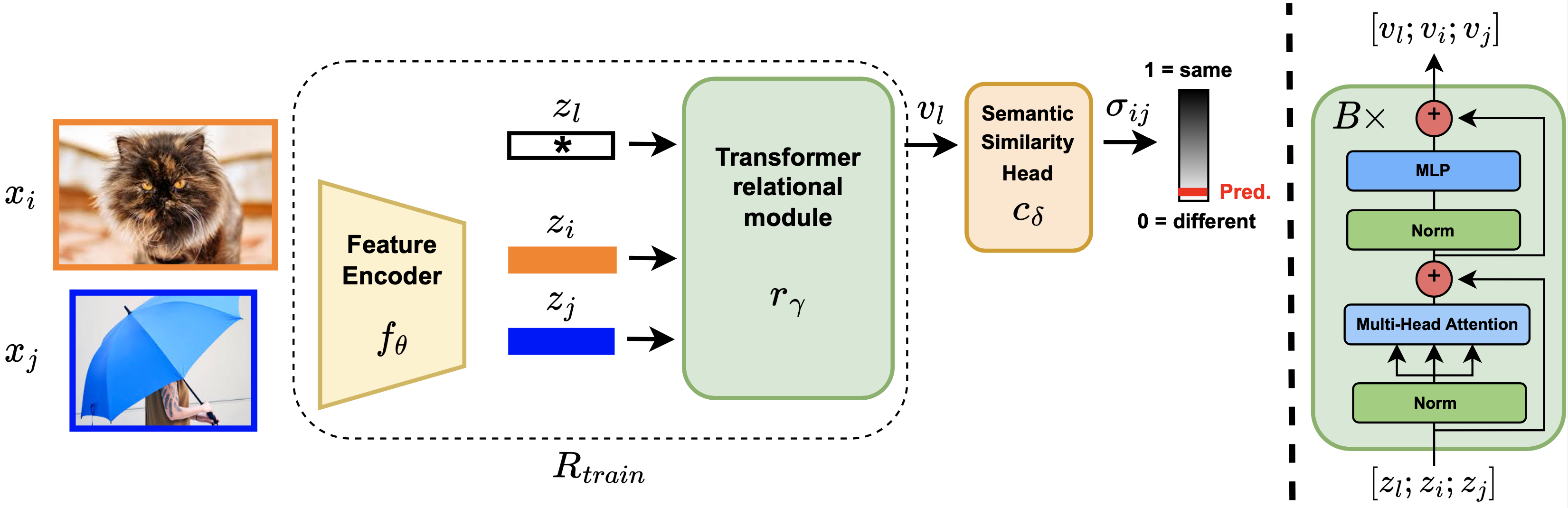}
    \caption{Schematic illustration of the training phase of \OUR. The features extracted from a pair of images are provided as input to our relational module. It consists of a transformer encoder that elaborates over a tuple composed of the sample pair and of a learnable label token. The output corresponding to this last token is finally provided as input to a semantic similarity head that predicts the sample resemblance.} 
    \label{fig:architecture}\vspace{-5mm}
\end{figure*}

\smallskip\noindent\emph{OOD strategies in literature}. %
From standard classification, we can evaluate whether a test sample is anomalous by applying a threshold on the output score of the top predicted class (maximum-softmax probability, MSP \cite{hendrycks17baseline}). Improvements over this basic approach have been proposed in \cite{liang2017enhancing,gODIN2020,liu2020energy}.
Instead of the model output, a recent work has shown how the gradients space of neural networks can be used to estimate prediction uncertainty and obtain an OOD scoring function \cite{huang2021importance}. 
Generative-based approaches consider the performance of a model trained on reference known classes when reconstructing an input sample. The reconstruction error defines the novelty score \cite{Kim2020RaPP,AE_anomaly_ICPR20,park2020learning}: GAN and flow-based invertible models have been exploited for this purpose \cite{aaai_SensoyKCS20,xia2020synthesize,nalisnick2018do}. Some methods synthesize out-of-distribution data \cite{Neal_2018_ECCV,GenOpenMax,lee2018training} or use external dataset as a source of {outlier exposure} during training  \cite{hendrycks2018deep,PAPADOPOULOS2021138,chen2021atom}.
\rev{A different solution consists in estimating test samples normality by computing their distance from training data using specific embeddings or metrics. \cite{NEURIPS2018_abdeb6f5,gram_ood}}.
A stream of works has also shown the effectiveness of self-supervised representation learning \cite{gidaris2018unsupervised,noroozi2016,Jenni_steering,yun2019cutmix,cutpaste}, and in particular of the contrastive-based strategies for OOD \cite{golan2018deep,Bergman2020Classification-Based,tack2020_CSI,sehwag2021_SSD,winkens2020contrastive}. 
Indeed by removing the focus from the labels, self-supervised models capture analogies and differences among the samples and provide a better way to score similarities. However, training these models %
needs a non-trivial optimization process with large training batches.
Embeddings based on self-attention have been considered as starting point for OOD in \cite{koner2021oodformer,fort2021_ExploringLimitsOOD}. Here the powerful transformer architecture ViT \cite{ViT_dosovitskiy2021an} pre-trained on ImageNet \cite{deng2009imagenet} for classification is finetuned on the training data to then score the test samples via MSP. Still, the risks of finetuning a large model on the training data for OOD were discussed in \cite{deeckeICML21}, which highlighted how part of the original knowledge gets lost in this process.

Finally, as also noticed by Huang et al. \cite{huang2021_MOS}, we underline how most of the existing works on OOD consider experimental analysis on datasets containing only digits or low-resolution images. Combined with the limitation of the existing models described above, it becomes clear the need for novel efficient solutions that can be easily deployed in real-world conditions.

\section{Method}
\subsection{Notation and background}

In the {semantic novelty detection} task, 
we have two datasets: 
a \emph{support set} containing labeled samples $\sS=\{\bx^{s},y^{s}\}_{k=1}^{K}$  drawn from the distribution $p_{\sS}$ and a \emph{test set} containing unlabeled samples $\sT = \{\bx^{t}\}_{h=1}^{H}$ drawn from the distribution $p_{\sT}$. 
The main difference between $p_{\sS}$ and $p_{\sT}$ is a semantic shift: it holds $y^{s}\in\sY_s$ and  $y^{t}\in\sY_t$, with $\sY_s \neq \sY_t$. The two sets of classes can be either completely disjoint $ \sY_s \cap \sY_t =  \emptyset$, or partially overlapping $\sY_s \subset \sY_t$. In the following we will indicate $\sY_s$ as the \emph{known} classes, while we use the term \emph{unknown} 
to refer to 
the test classes $\sY_{t \setminus s}$ not appearing in the support set. 
Domain shift may contribute to the distribution difference among support and test, causing a variation in the appearance of the samples. Still, the class content remains unchanged. 
A reliable semantic novelty detector should discriminate between known and unknown samples in the test set while being robust to the domain shift.

Given a test sample $\bx^t$, the detector $D$ should be able to predict a $score \in [0,1]$ that signals whether it is known or unknown with respectively high and low values. Following the traditional strategy, the detector can be formalized as  $D:\{C_{train}(\sI), N_{train}(\sS), N_{eval}(\bx^t)\}$. 
At first a good representation is learned by training a classification model $C$ on the samples $(\bx_i, y_i)_{i=1}^I$ of a large-scale dataset $\sI$ as ImageNet1k \cite{deng2009imagenet}. The representation is then inherited by the model $N$ which is finetuned on the support set to gather the definition of \emph{normality} from the data. When this training is guided by a simple classification objective, the final evaluation of $N$ on the test is usually performed by MSP: $score = \max_{c\in\sY_s}p(y=c|\bx^t)$.
We highlight that the finetuning process has a computational cost that might not be affordable on edge devices. Moreover, in the long term, its catastrophic forgetting effect reduces the original large-scale knowledge, as well as the ability to anticipate potential semantic anomalies \cite{deeckeICML21}. Thus, carefully designing the representation learning approach and choosing how the pre-trained model should be applied for the downstream task is crucial. %

We propose to change the learning paradigm for the semantic novelty detector so that it can be written as $D:\{R_{train}(\sI), N_{eval}(\sS,\bx^t)\}$. The first component $R$ is a representation learning model based on relational reasoning and trained on ImageNet1k. The learned embedding is directly used by an evaluation system to compare each test sample with the 
support set to obtain its normality score.

\subsection{Representation Learning via Relational Reasoning}
\label{sec:method_loss}
We consider $R$ composed of %
a feature extractor $f_\theta$ and a relational module $r_\gamma$. 
A pair of samples $(\bx_i,\bx_j)$ from the reference dataset $\sI$ passes first through the feature extractor $(\bz_i = f_\theta(\bx_i),\bz_j = f_\theta(\bx_j))$, and is then fed to the relational module $r_\gamma$.
The output of this module is the input of the semantic similarity head $c_\delta$ which is simply a fully connected (FC) layer. It returns $\sigma_{ij} = c_\delta(r_\gamma(\bz_i, \bz_j)) \in [0,1]$ which represents a semantic similarity measure and can be interpreted as the probability that the two input samples belong to the same category.

The whole representation learning model is trained with a regression objective. Specifically, we assign to each data pair the label $l_{ij}=0$ if $y_i \neq y_j$ and $l_{ij}=1$ otherwise, and we minimize the MSE loss: 
\begin{align}
\label{eq:loss}
    \argmin_{\theta, \gamma, \delta} \sum_{m=1}^M (\sigma_m - l_m)^2~,
\end{align}
\rev{here the index $m$ specifies the pairs $(\bx_i,\bx_j)$ with $i\neq j$ and $x_i,x_j \in \sI$}. \\
\noindent Despite the ground truth supervision being only at the extremes of the prediction interval, we aim at learning a semantic similarity measure in the continuous range $[0,1]$. For this reason, the regression loss is particularly suited for the task, but the problem could be also casted as binary classification. In the experimental section we compare the two approaches providing empirical evidences about the beneficial effect of our regression choice.

\subsection{Evaluation Process}
Starting from the learned embedding, the component $N_{eval}$ of our approach has the simple role of comparing each test sample with the reference support set, without any further training phase.
$N_{eval}$ exploits the relational module and provides to it data pairs composed of the feature of each test sample $\bz^t=f_\theta(\bx^t)$, and the set of per-class prototypes %
$\overline{\bz}^s_{y^s}~ \forall ~y^s\in\sY_s$ obtained as the average over the samples of each class in the support set.
We obtain a vector $\bu$ of  $|\sY_s|$ elements, each corresponding to 
$c_\delta(r_\gamma(\bz^t,\overline{\bz}^s_{y^s}))$ and expressing the similarity of the test sample $\bz^t$ to one of the known classes. 
This output is filtered by a softmax function and we apply MSP to get the final normality score: $score = \max(softmax(\bu))$.

\subsection{Relational module}
With respect to other standard components of deep neural networks that elaborate on single samples, the peculiarity of the relational module is that it processes pairs of inputs to provide information on their similarity. Of course, the order of appearance of the two samples should not influence the network output as any good similarity measure needs to be symmetric.
Considering its natural permutation invariance and its well known capability of comparing multiple inputs, we implement our relational module through a simple transformer encoder.
It consists of $B$ identical blocks, each one composed of a Multi-Head Self-Attention (MSA) and a Multi-Layer Perceptron (MLP), both preceded by Layer-Norm (LN) modules and bypassed by residual skip connections as shown in the right part of Fig. \ref{fig:architecture}.
The input feature vectors pair, together with a \rev{learnable} label token, forms the tuple $[\bz_{l},\bz_i,\bz_j]$ which is fed as input to the transformer and passes through all its layers, \rev{producing the output sequence $[\bv_{l},\bv_i,\bv_j]$}. Note that, in this architecture each image represents a single input token to the transformer, as done in \cite{Cheng2020_AudioVisualTransf}.
We do not include in our encoder the commonly used positional embeddings as we aim at keeping the permutation invariance.
In our implementations we use a ResNet18-based backbone as feature extractor $f_\theta$ and select $\bv_{l}$ as the output of the relational module $r_\gamma$, \rev{which is then passed through the head $c_{\delta}$ to produce a semantic similarity score $\sigma_{ij}$}. %
\rev{In the experimental section we evaluate alternative architectures for our relational module.}

\section{Experimental Setup}
With \OUR we are proposing a novel strategy fully based on representation learning for OOD. We claim that the embedding space learned via relational reasoning is well suited to detect novel classes simply comparing the test samples with the support set which represents the \emph{normal} reference condition. Since this logic substantially differs from that of previous works in OOD, %
there are several questions that we need to answer with experimental validations. 

\smallskip\noindent\emph{Are existing representation learning approaches effective for the SeND task?} (see Sec. \ref{ssec:intra-domain})
We focus on the data representation learned via a pre-training stage on ImageNet1k.
We consider several state-of-the-art learning methods and for all of them, we keep the same prototype-based evaluation strategy used for \OUR: every class of the support set is identified by averaging on the feature representation of its samples, and the normality score for each test instance is evaluated by measuring the similarity with the nearest known class centroid. 

We choose two families of methods.
Among the cross-entropy based classifiers we consider 
this loss %
applied to \textbf{ResNet} \cite{he2016deep}  and \textbf{ViT} \cite{ViT_dosovitskiy2021an} architectures, and the data augmentation-based approach \textbf{CutMix} \cite{yun2019cutmix}. 
For the contrastive learning techniques we consider the self-supervised methods \textbf{SimCLR} \cite{simclr2020} and \textbf{CSI} \cite{tack2020_CSI}, as well as their supervised versions \textbf{SupCLR} \cite{khosla2020supervised} and \textbf{SupCSI} \cite{tack2020_CSI}. 
The relation between each test sample and the class prototypes is measured via the Euclidean similarity (inverse of the Euclidean distance \cite{Segaran07}) and the cosine similarity, respective for the cross-entropy and contrastive approaches.
We highlight that these methods appeared before in the anomaly and novelty detection literature \cite{cutpaste,koner2021oodformer,anomaly_detection_survey_dietterich}, but their application always involved a training phase on the support set, while here we run them only on ImageNet1k to get their learned representation. 
Note also that the different names identify the characteristics of their learning objective, but all of them share the same backbone architecture: ResNet101 \cite{he2016deep} with 44M learnable parameters, comparable to the 40M of \OUR (11M for $f_\theta$, 29M for $r_\gamma+c_\delta$). The only exception is ViT, that we included as an example of Vision Transformer whose usage for OOD was suggested in \cite{koner2021oodformer}, and for which we use the Vit-Base (86M \rev{parameters}) implementation from \cite{DeiT_pmlr-v139-touvron21a}. 

\smallskip\noindent\emph{Is the learned embedding robust to domain variations?} (See Sec. \ref{ssec:cross-domain})
ImageNet1k contains pictures of real-world objects and it is important to check if the relations encoded in the learned embedding are still relevant when the final goal is to identify novel classes in completely different contexts as for texture images or among sketches. We consider two levels of difficulty. The first is due to a domain difference between the pre-training and the downstream task: the support and test set are drawn from the same domain which is different from that of ImageNet1k. 
The second is a domain generalization problem and consider also a domain shift between the support and the test set. The support set can be composed of data from a single or multiple domain sources, while the test is from a target domain. %
\rev{We exploits several datasets to perform a thorough analysis.}

\smallskip\noindent \textbf{Textures} \cite{Cimpoi_2014_CVPR_textures} is a collection of textural images, it consists of 5,640 images organized in 47 categories. We randomly chose 23 categories as known and 24 as unknown.
\textbf{DomainNet} \cite{peng2019moment_domainnet} is a large-scale dataset of common objects from six different domains with 345 object categories. We use this dataset for both intra-domain and cross-domain experiments. For the first case,  we used the Natural Language Toolkit \cite{bird2009natural} to select 50 categories that do not overlap with ImageNet1k classes. We then randomly selected 25 as known and 25 as unknown.
\textbf{PACS} \cite{li_2017_pacs} is composed of four domains and 7 object categories. We follow the known/unknown division proposed in \cite{shu2021_OPENDG} using 6 categories as known and 1 as unknown.
\textbf{OfficeHome} \cite{venkateswara2017deep} consists of four domains and 65 categories. We use it in the single-domain generalization experiments by following \cite{bucci2020ros} for the known/unknown category division (25 known and 40 unknown categories). 
We adopt the same setting of \cite{shu2021_OPENDG} for the multi-source cross-domain experiments.
\textbf{Multi-Datasets} is a very realistic setting proposed in \cite{shu2021_OPENDG} where the multi-source condition is naturally determined by the use of several datasets as source domains: {Office-31} \cite{saenko2010adapting}, {STL-10} \cite{coates2011analysis}, {Visda2017} \cite{peng2017visda}. The partial overlap between the source categories, that is simulated for the PACS and OfficeHome benchmarks, in this case is naturally obtained. Here the target domains (Clipart, Real, Painting, Sketch) come from DomainNet.

\smallskip\noindent\emph{How does \OUR compare with state-of-the-art OOD methods?} (See Sec. \ref{ssec:finetuning})
Considering that \OUR does not need access to the support set in the training stage but relies on it during the evaluation, we can measure the time and computational resources it uses in this last stage and provide the same to the training procedure of state-of-the-art OOD methods. 
We consider the following baselines: \textbf{MSP} \cite{hendrycks17baseline} which uses the standard maximum softmax probability, \textbf{ODIN} \cite{liang2017enhancing} a simple approach based on input perturbation and temperature scaling, \textbf{Energy} \cite{liu2020energy} that uses an energy score for OOD uncertainty estimation, \textbf{GradNorm} \cite{huang2021importance} which relies on test-time extracted gradients to detect the out-of-distribution samples, the ViT-based approach \textbf{OODFormer} \cite{koner2021oodformer} \rev{and two methods based on tailored metric estimation: Mahalanobis \cite{NEURIPS2018_abdeb6f5} and Gram \cite{gram_ood}}.

\smallskip\noindent\emph{Can \OUR provide unknown detection abilities to closed-set approaches?} (See Sec. \ref{ssec:os-dg})
\OUR does not need any training on the support set and it may work as a plug-and-play module to provide close-set approaches the ability to work in open-set conditions. 
\rev{We focus on the challenging open-set domain generalization (DG) setting presented in \cite{shu2021_OPENDG} and show how \OUR can enhance existing approaches. Besides \textbf{DAML} introduced in \cite{shu2021_OPENDG}, we consider the state-of-the-art multi-source closed-set DG method \textbf{SWAD} \cite{cha2021swad}, which looks for flat minima in the learning objective function, and two single-source closed-set methods: \textbf{SagNet} \cite{nam2021reducing} disentangles shape from style in the image features to reduce the style bias, while \textbf{Diversify} \cite{wang2021learning} synthesizes images with unseen styles.}

\section{Experiments}
\begin{table}[t]
\centering
\caption{Intra-Domain analysis. Best result in bold and second best underlined.}
\resizebox{\textwidth}{!}{
\begin{tabular}{|c|c|c@{~~}c|c@{~~}c|c@{~~}c|c@{~~}c|}
\hline
\multicolumn{1}{|c}{\multirow{2}{*}{\small{Rep. Learning}}} & \multicolumn{1}{|c|}{\multirow{2}{*}{\small{Network}}} 
&  \multicolumn{2}{c|}{\small{Texture}} & \multicolumn{2}{c|}{\small{Real}} &  \multicolumn{2}{c|}{\small{Sketch}}  & \multicolumn{2}{c|}{\small{Painting}}  \\
 & & \scriptsize{AUROC} $\uparrow$ & \scriptsize{FPR95} $\downarrow$ &  \scriptsize{AUROC} $\uparrow$ & \scriptsize{FPR95} $\downarrow$ &  \scriptsize{AUROC} $\uparrow$ & \scriptsize{FPR95} $\downarrow$  &  \scriptsize{AUROC} $\uparrow$ & \scriptsize{FPR95} $\downarrow$   \\
\hline
\small{Cross Entropy }  & \small{ResNet \cite{he2016deep}}  & \underline{0.678} & \underline{0.892} & {0.710} & {0.860} & {0.553}	 & \underline{0.936} & 0.651 &	0.926  \\
\small{Cross Entropy }  &\small{ViT \cite{ViT_dosovitskiy2021an}}  & 0.562	 & 0.919  & 0.696  &	\underline{0.833} & \underline{0.554} &	0.952  & \underline{0.681}	& \underline{0.850}  \\
 \small{CutMix \cite{yun2019cutmix}}  & \small{ResNet}  & 0.619 & 0.922 & \underline{0.721}  & 0.877	 & 0.542  &	{0.943}  &   0.629 & 0.927	\\
\small{SimCLR \cite{simclr2020}} & \small{ResNet}  
 & 0.529 &	0.942  & 0.481 &	0.944 & 0.502 &	0.956  & 0.510 &	0.956 \\
\small{SupCLR \cite{khosla2020supervised}} & \small{ResNet}   & 0.534	 &  0.947 &  0.561 &	0.899  &  0.532 &	0.946  & 0.532 &	0.933  \\
\small{CSI \cite{tack2020_CSI}} & \small{ResNet}
 & 0.651 &	0.906  &  0.663 &	0.887 &  0.514	 & 0.955 &  0.621 &	0.910 \\
\small{SupCSI \cite{tack2020_CSI}} & \small{ResNet}
  & {0.652} &	 {0.903}  &  0.695 &	0.875 &  0.535 &	0.953  &   {0.652} &	 {0.909}    \\
\hline
\hline
 \multicolumn{2}{|c|}{\small{\textbf{\OUR}}} & \textbf{0.691} &	\textbf{0.859} &	\textbf{0.780} &	\textbf{0.805} &	\textbf{0.623} &	\textbf{0.917} &	\textbf{0.735} &	\textbf{0.829}\\
\hline
\end{tabular}
} 
\label{tab:singledomain_new}
\end{table}
Here we report and discuss the results of our experimental analysis. %
All the evaluations are done on the basis of two metrics. \textbf{AUROC} is the Area Under the Receiver Operating Characteristic curve, obtained by varying the normality decision threshold. \textbf{FPR95} corresponds to the false positive rate of out-of-distribution examples when the true positive rate of in-distribution examples is at 95\%. \rev{For the open-set DG experiments we follow \cite{shu2021_OPENDG} and consider also the overall accuracy on the known samples \textbf{Acc} and the harmonic mean between the accuracy on known classes and the unknown detection accuracy \textbf{H-score}.}
Implementation\footnote{\rev{The code is available at \url{https://github.com/FrancescoCappio/ReSeND}}} details and more experimental analyses are provided in the Appendices \ref{ref:app_imp_details} and \ref{ref:app_results}. All experimental results are averaged over three runs.

\subsection{Intra-Domain analysis} 
\label{ssec:intra-domain}
For the intra-domain analysis, we consider the support and test sets drawn from the same visual distribution but showing significant differences from ImageNet1k. In particular, all the testbeds were explicitly designed to avoid semantic overlaps with ImageNet1k: this means that neither known nor unknown classes appear in its label set. Variation in data type and domain further enlarge the appearance gap.
The texture benchmark \cite{Cimpoi_2014_CVPR_textures} was already used in \cite{huang2021_MOS} and covers a completely different data type with respect to ImageNet1k  (objects vs textures). 
Real, Sketch and Painting benchmarks are obtained from the DomainNet dataset \cite{peng2019moment_domainnet} and, differently from Texture, they share the same data type (objects) of ImageNet1k and cover the same (Real) or different (Sketch, Painting) visual domains. In Table \ref{tab:singledomain_new} we can see that \OUR obtains the best results showing an excellent knowledge transfer capability. On Texture, the second and third best are respectively Cross Entropy on ResNet and SupCSI, but this ranking is not consistent over all the settings and the performance gap with respect to \OUR  remains evident, especially in the case of Sketch and Painting. 

\subsection{Cross-Domain analysis}
\label{ssec:cross-domain}
In many real-world conditions, it's impossible to avoid the presence of a visual domain shift between training and test data. This usually increases the complexity of the task at hand. A reliable semantic novelty detection method should disregard the domain shift between the support and the test set, focusing only on the semantic content of the data. 
We compare \OUR with the same baselines of the previous section, considering two different benchmarks built from the PACS dataset \cite{li_2017_pacs}. Here the support set is composed of images of the source domain, while the target domain is used as test set. In the single-source case (Table \ref{tab:crossdomain_analysis_new} top), the Photo domain is always used as source, while the three remaining domains are used as target. The multi-source benchmark (Table  \ref{tab:crossdomain_analysis_new} bottom) is inherited from \cite{shu2021_OPENDG}: each domain is used in turn as target, with the additional difficulty that the support set is composed by multiple sources that have a partial class overlap (see Fig. \ref{fig:opendg_setting}). 
We notice that SimCLR is particularly effective when the test domain is sketch, but it is outperformed by other approaches in the remaining settings. On the other hand, 
\OUR is able to obtain top results in all benchmarks, showing high robustness to the domain shift, despite not including any tailored strategy designed for bridging it. 

\begin{table*}[t]
\centering
\caption{Cross-domain analysis. Top: single-source results, Bottom: multi-source results. We consider the PACS dataset with all the possible combinations of source/target as support/test sets. %
Best result in bold and second best underlined.}
\label{tab:crossdomain_analysis_new}
\resizebox{\textwidth}{!}{
\begin{tabular}{|c|c|c@{~~}c|c@{~~}c|c@{~~}c||c@{~~}c|}
\hline
\multicolumn{1}{|c}{\multirow{3}{*}{\small{Rep. Learning}}} & \multicolumn{1}{|c|}{\multirow{3}{*}{\small{Network}}} 
&  \multicolumn{8}{c|}{\small{PACS \textbf{Single-Source}}} \\
\cline{3-10}
& & \multicolumn{2}{c|}{{\small{ArtPainting}}}  & \multicolumn{2}{c|}{{\small{Sketch}}}  & \multicolumn{2}{c||}{{\small{Cartoon}}}  & \multicolumn{2}{c|}{{\small{\textbf{Avg}}}} \\
&  & \scriptsize{AUROC} $\uparrow$ & \scriptsize{FPR95} $\downarrow$ &  \scriptsize{AUROC} $\uparrow$ & \scriptsize{FPR95} $\downarrow$ & \scriptsize{AUROC} $\uparrow$ & \scriptsize{FPR95} $\downarrow$ & \scriptsize{AUROC} $\uparrow$ & \scriptsize{FPR95} $\downarrow$   \\
\hline
Cross Entropy& \multicolumn{1}{c|}{\small{ResNet \cite{he2016deep}}} & 0.655  & 	0.940 &	0.519 &	0.969 &	0.546 &	0.958 &	0.573 &	0.956  	\\
Cross Entropy& \multicolumn{1}{c|}{\small{ViT \cite{ViT_dosovitskiy2021an}}} &  0.593 &	0.895 &	0.595 &	0.881 &	0.500 &	0.953 &	0.562 &	0.910 	 \\
 \multicolumn{1}{|c}{\small{CutMix \cite{yun2019cutmix}}}&  \multicolumn{1}{|c|}{\small{ResNet}} &  0.663 & 0.949 & 0.372 & 0.981 & 0.419 & 0.980 & 0.485 & 0.970 \\
\multicolumn{1}{|c}{\small{SimCLR \cite{simclr2020}}} & \multicolumn{1}{|c|}{\small{ResNet}}
 & 0.444 &	0.984 &	\textbf{0.945} &	\textbf{0.400} &	0.401 &	0.988 &	\underline{0.597} &	\textbf{0.791}	\\
\multicolumn{1}{|c}{\multirow{1}{*}{\small{SupCLR \cite{khosla2020supervised}}}} & \multicolumn{1}{|c|}{\small{ResNet}} & 0.500  &	0.909 &	0.176 &	1.000 &	0.469 &	0.919 &	0.381 &	0.942 	\\
\multicolumn{1}{|c}{\small{CSI \cite{tack2020_CSI}} } & \multicolumn{1}{|c|}{\small{ResNet}}
 & 0.495 &	0.987 &	0.591 &	0.881 &	0.433 &	0.978 &	0.506 &	0.949	 \\
\multicolumn{1}{|c}{\small{SupCSI \cite{tack2020_CSI}} } & \multicolumn{1}{|c|}{\small{ResNet}}
  &  0.546 &	0.976 &	\underline{0.655} &	\underline{0.819} &	\underline{0.567} &	\underline{0.909} &	0.589 &	0.901	\\
\hline
\hline
\multicolumn{2}{|c|}{\multirow{1}{*}{\small{\textbf{\OUR}}}}   & \textbf{0.828} &	\textbf{0.668} &	0.576 &	0.981 &	\textbf{0.651} &	\textbf{0.891} &	\textbf{0.685} &	\underline{0.847}	 \\
\hline
\end{tabular}
}\vspace{1mm}
\resizebox{\textwidth}{!}{
\begin{tabular}{|c|c|c@{~~}c|c@{~~}c|c@{~~}c|c@{~~}c||c@{~~}c|}
\hline
\multicolumn{1}{|c|}{\multirow{3}{*}{\small{Rep. Learning}}} & \multicolumn{1}{c}{\multirow{3}{*}{\small{Network}}} 
&  \multicolumn{10}{|c|}{\small{PACS \textbf{Multi-Source}}} \\
\cline{3-12} 
 & &  \multicolumn{2}{c|}{{\small{ArtPainting}}}  & \multicolumn{2}{c|}{{\small{Sketch}}}  & \multicolumn{2}{c|}{{\small{Cartoon}}}  & \multicolumn{2}{c||}{{\small{Photo}}} & \multicolumn{2}{c|}{{\small{\textbf{Avg}}}}  \\
 & &  \scriptsize{AUROC} $\uparrow$ & \scriptsize{FPR95} $\downarrow$ &  \scriptsize{AUROC} $\uparrow$ & \scriptsize{FPR95} $\downarrow$ & \scriptsize{AUROC} $\uparrow$ & \scriptsize{FPR95} $\downarrow$ & \scriptsize{AUROC} $\uparrow$ & \scriptsize{FPR95} $\downarrow$ & \scriptsize{AUROC} $\uparrow$ & \scriptsize{FPR95} $\downarrow$  \\
\hline
Cross Entropy & \multicolumn{1}{c|}{\small{ResNet \cite{he2016deep}}}  & 0.575 &	0.947 &	0.451 &	1.000	& 0.547 &	0.943 & 0.361 &	0.991 & 0.484 &	0.970	\\

Cross Entropy &\multicolumn{1}{c|}{\multirow{1}{*}{\small{ViT \cite{ViT_dosovitskiy2021an}}}}   & \underline{0.611} &	\underline{0.837} &	0.566 &	0.944  & 0.539 &	0.904	& \underline{0.932}	 & \underline{0.403}	 & \underline{0.662} &	\underline{0.772}	 \\

\multicolumn{1}{|c|}{\small{CutMix \cite{yun2019cutmix}}}& \multicolumn{1}{c|}{\small{ResNet}} & 0.604 & 0.895 & 0.411 & 1.000 & 0.407 & 0.975 & 0.655 & 0.942 & 0.519 & 0.953   \\

\multicolumn{1}{|c|}{\small{SimCLR \cite{simclr2020}}} &\multicolumn{1}{c|}{\small{ResNet}}
 &	 0.461 &	0.953	 & \textbf{0.933} &	\textbf{0.663} & 0.368 &	0.995  &  0.739	 & 0.854 & 0.625 &	0.866	\\
\multicolumn{1}{|c|}{\small{SupCLR \cite{khosla2020supervised}}}& \multicolumn{1}{c|}{\small{ResNet}} & 0.581 &	0.898 &	0.100 &	1.000 & 0.499 &	0.909	 & 0.467 &	0.995	 & 0.412 &	0.951		\\
\multicolumn{1}{|c|}{\small{CSI \cite{tack2020_CSI}}}&\multicolumn{1}{c|}{\small{ResNet}}
&	0.474 &	0.984 &	\underline{0.702} &	\underline{0.800} &	\underline{0.560} &	\underline{0.977} &	0.524 &	0.946 &	0.565 &	0.927	 \\
\multicolumn{1}{|c|}{\small{SupCSI \cite{tack2020_CSI}}} &\multicolumn{1}{c|}{\small{ResNet}}
 & 0.417	& 0.984	 & 0.660 &	0.869 &	0.323 &	1.000 &	0.601 &	0.946 &	0.500 &	0.950	\\
\hline
\hline
\multicolumn{2}{|c|}{\multirow{1}{*}{\small{\textbf{\OUR}}}}  & \textbf{0.750} &	\textbf{0.820} &	0.685	 & 0.894 & \textbf{0.660} &	\textbf{0.854} & \textbf{0.963} &	\textbf{0.181} & \textbf{0.765} &	\textbf{0.687}	 \\
\hline
\end{tabular} 
}
\end{table*}
\begin{table*}[t]
\centering
\caption{Comparison with finetuning-based state-of-the-art OOD methods. 
Best result in bold and second best underlined.}
\resizebox{\textwidth}{!}{
\begin{tabular}{|cc|cc|c@{~~}c|c@{~~}c|c@{~~}c|c@{~~}c||c@{~~}c|}
\hline
\multicolumn{2}{|c|}{\multirow{3}{*}{\small{OOD Methods}}} 
 & & \multicolumn{10}{c}{\small{PACS \textbf{Multi-Source}}} & \multicolumn{1}{c|}{} \\
\cline{3-14} 
& &  \multicolumn{1}{c|}{\multirow{2}{*}{Fine-Tun.}} & \multirow{2}{*}{Eval.} & \multicolumn{2}{c|}{{\small{ArtPainting}}}  & \multicolumn{2}{c|}{{\small{Sketch}}}  & \multicolumn{2}{c|}{{\small{Cartoon}}}  & \multicolumn{2}{c||}{{\small{Photo}}} & \multicolumn{2}{c|}{{\small{\textbf{Avg}}}}  \\
 & & \multicolumn{1}{c|}{} & & \scriptsize{AUROC} $\uparrow$ & \scriptsize{FPR95} $\downarrow$ &  \scriptsize{AUROC} $\uparrow$ & \scriptsize{FPR95} $\downarrow$ & \scriptsize{AUROC} $\uparrow$ & \scriptsize{FPR95} $\downarrow$ & \scriptsize{AUROC} $\uparrow$ & \scriptsize{FPR95} $\downarrow$ & \scriptsize{AUROC} $\uparrow$ & \scriptsize{FPR95} $\downarrow$  \\
\hline
\multicolumn{2}{|c|}{MSP \cite{hendrycks17baseline} }  & \multicolumn{1}{c|}{\checkmark } & \checkmark  & 0.617 &	0.973 & 0.412 &	0.998 & \underline{0.781} &	\textbf{0.767} & 0.752 &	0.905 &	0.640 &	0.911  \\
\multicolumn{2}{|c|}{ODIN \cite{liang2017enhancing} }  & \multicolumn{1}{c|}{\checkmark } & \checkmark  & 0.602 &	0.977 & 0.425 &	0.998 & \textbf{0.785} &	\underline{0.774} & 0.782 &	0.912 &	0.649 &	0.915 \\
\multicolumn{2}{|c|}{Energy \cite{liu2020energy} }  & \multicolumn{1}{c|}{\checkmark } & \checkmark  & 0.583 & 0.987  & 0.543 &	0.996 & 0.687 &	0.802 & 0.845 &	0.924 &	0.665 &	0.927\\
\multicolumn{2}{|c|}{GradNorm \cite{huang2021importance} }  & \multicolumn{1}{c|}{\checkmark } & \checkmark  & 0.637 &	0.954  & 0.514 &	1.000 & 0.762 &	\textbf{0.767} & 0.851 &	0.861 &	0.691 &	0.896\\
\multicolumn{2}{|c|}{OODformer \cite{koner2021oodformer} }  & \multicolumn{1}{c|}{\checkmark } & \checkmark  & \underline{0.703} &	\underline{0.929} & 0.610 &	0.973 & 0.776 &	0.802 & 0.732 &	0.773 &	\underline{0.705} &	\underline{0.869}\\
\multicolumn{2}{|c|}{\rev{Mahalanobis} \cite{NEURIPS2018_abdeb6f5} }  & \multicolumn{1}{c|}{\checkmark } & \checkmark  & 0.596 &	0.976 & 0.559 &	0.933 & 0.682 &	0.909 & \underline{0.861} &	0.849 &	0.665 &	0.916 \\
\multicolumn{2}{|c|}{\rev{Gram} \cite{gram_ood} }  & \multicolumn{1}{c|}{\checkmark } & \checkmark  & 0.448 &	0.962 & \textbf{0.885} &	\textbf{0.713} & 0.536 &	0.946 & 0.838 &	\underline{0.579} &	0.677 &	0.800\\

\hline
\hline
\multicolumn{2}{|c|}{\multirow{1}{*}{\small{\rev{Mahalanobis} \cite{NEURIPS2018_abdeb6f5}}}} & \multicolumn{1}{c|}{\ding{53} } & \checkmark   & 0.596 &	0.976 &	0.466	 & 0.981 & 0.593	 & 0.926 & 0.808 & 0.935 & 0.616 &	0.954 \\
\multicolumn{2}{|c|}{\multirow{1}{*}{\small{\rev{Gram} \cite{gram_ood}}}} & \multicolumn{1}{c|}{\ding{53} } & \checkmark   & 0.494 &	0.960 &	\underline{0.840} & \underline{0.844} & 0.494 &	0.954 & 0.797 &	0.981 & 0.656 &	0.935	 \\
\multicolumn{2}{|c|}{\multirow{1}{*}{\small{\textbf{\OUR}}}} & \multicolumn{1}{c|}{\ding{53} } & \checkmark   & \textbf{0.750} &	\textbf{0.820} &	0.685	 & 0.894 & 0.660 &	0.854 & \textbf{0.963} &	\textbf{0.181} & \textbf{0.765} &	\textbf{0.687}	 \\
\hline
\end{tabular}
} 
\label{tab:finetuning_new} 
\end{table*}

\subsection{OOD with budget-limited finetuning}
\label{ssec:finetuning}
As previously discussed, \OUR doesn't need finetuning on the support set to be used for semantic novelty detection. Hence it is not trivial to make a fair comparison with existing OOD methods for which instead the learning phase on the support set is essential. Nevertheless, we believe that it's important to contextualize \OUR in the current literature to provide a clearer overview of its performance. With this objective in mind, we focus on the challenging PACS multi-source setting and compare against a number of standard and state-of-the-art OOD methods by letting them learn (refine the original ImageNet1k pretrained model) on the support set for the same time and using the same computational resources exploited by \OUR in the prediction phase ($\sim30$s on 1 GPU for the considered benchmark). \rev{For what concerns Mahalanobis \cite{NEURIPS2018_abdeb6f5} and Gram \cite{gram_ood}, given that they are metric-based methods, the distance between test samples and the support set can be computed also using a non-finetuned model %
(although this was not the strategy proposed by the authors). Thus,  
we tested both the finetuned and not finetuned versions.}
The results in Table \ref{tab:finetuning_new} show that \OUR clearly outperforms all the competitors, which would need a longer training period or more resources in order to converge to a good model. This confirms the role of \OUR as a powerful tool when semantic novelty detection is performed under restrictive budget constraints. 

\subsection{Open-set Domain Generalization}
\label{ssec:os-dg}
\begin{figure}[t]
\begin{minipage}[]{.35\textwidth}
\vspace{4.5mm}
\begin{figure}[H]
	\includegraphics[width=0.93\linewidth]{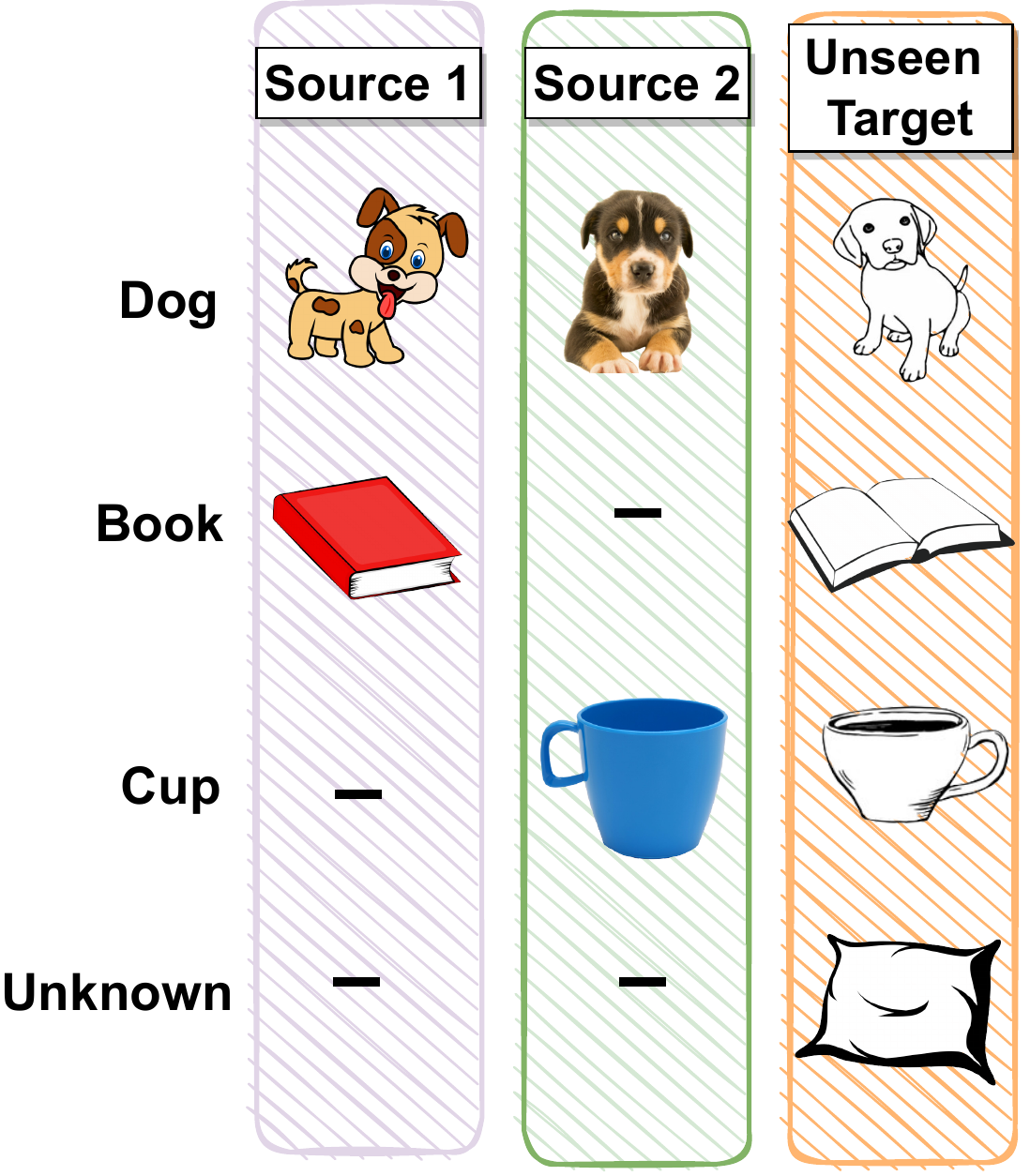}
	\vspace{-4mm}
	\caption{Open-Set DG setting\label{fig:opendg_setting}}
\end{figure}
\end{minipage}
	\begin{minipage}[]{.65\textwidth}
		\begin{table}[H]
			\caption{Open-Set DG experiments.}
			
			\resizebox{\linewidth}{!}{
				\begin{tabular}{c|ccc ccc ccc|}
					\cline{2-10}
					& \multicolumn{9}{c|}{\multirow{1}{*}{\small{\textbf{Single-Source}}}}  \\
					\cline{2-10}
					\multicolumn{1}{c|}{} & &   \multicolumn{3}{c|}{\small{PACS}} & \multicolumn{3}{c}{\small{Office-Home}} & &  \multicolumn{1}{c|}{}\\
					\multicolumn{1}{c|}{} & &   {\small{AUROC}}  & {\small{Acc}}  & \multicolumn{1}{c|}{{\small{H-Score}}} &   {\small{AUROC}}  & {\small{Acc}}  & {\small{H-Score}} & &  \multicolumn{1}{c|}{} \\
					\hline
					\multicolumn{1}{|c|}{{\small{\textbf{\OUR}}}} & &  0.685 & - &  \multicolumn{1}{c|}{-} & 0.685 & - & -  &  &   \multicolumn{1}{c|}{}\\
					\hline
					
					\multicolumn{1}{|c|}{{\small{SagNet \cite{nam2021reducing} + MSP}}} &  & 0.643 &	55.85 &	\multicolumn{1}{c|}{48.64}  & 0.699 &	67.58 &	59.92 & & \multicolumn{1}{c|}{} \\
					
					\multicolumn{1}{|c|}{{\small{SagNet+ \textbf{\OUR}}}} & & \textbf{0.700} &	55.85 &	\multicolumn{1}{c|}{\textbf{52.17}} & \textbf{0.714} &	67.58 &	\textbf{61.01} & & \multicolumn{1}{c|}{} \\
					\hline
					
					\multicolumn{1}{|c|}{{\small{Diversify \cite{wang2021learning} + MSP}}} & &  0.643	 & 52.06 &	\multicolumn{1}{c|}{48.12} & 0.696	& 70.49 &	60.03 & & \multicolumn{1}{c|}{} \\
					
					\multicolumn{1}{|c|}{{\small{Diversify+ \textbf{\OUR}}}} & &  \textbf{0.691} &	52.06 &	\multicolumn{1}{c|}{\textbf{51.19}} &  \textbf{0.707} &	70.49 &	\textbf{60.77} & &  \multicolumn{1}{c|}{} \\ 
					\hline
				\end{tabular}
			}			
			\resizebox{\linewidth}{!}{
				\begin{tabular}{c|ccc ccc ccc|}
					\cline{2-10}
					& \multicolumn{9}{c|}{\multirow{1}{*}{\small{\textbf{Multi-Source}}}}  \\
					\cline{2-10}
					&  \multicolumn{3}{c|}{\small{PACS}} & \multicolumn{3}{c|}{\small{Office-Home}} & \multicolumn{3}{c|}{\small{Multi-Datasets}}  \\
					
					&  {\small{AUROC}}  & {\small{Acc}}  & \multicolumn{1}{c|}{{\small{H-Score}}}  &  {\small{AUROC}}  & {\small{Acc}}  & \multicolumn{1}{c|}{{\small{H-Score}}} &  {\small{AUROC}}  & {\small{Acc}}  & {\small{H-Score}} \\
					\hline
					\multicolumn{1}{|c|}{{\small{\textbf{\OUR}}}}& 0.765 & - & \multicolumn{1}{c|}{-} & 0.674 & - & \multicolumn{1}{c|}{-} & 0.686 & - & - \\
					\hline
					
					\multicolumn{1}{|c|}{{\small{DAML \cite{shu2021_OPENDG} + MSP}}} & 0.657  & 62.85 & \multicolumn{1}{c|}{52.99} & 0.651 & 55.28 & \multicolumn{1}{c|}{52.37} & 0.695 &	45.90 &	47.88 \\
					\multicolumn{1}{|c|}{{\small{DAML+\textbf{\OUR}}}}& \textbf{0.722} &	62.85 &	\multicolumn{1}{c|}{\textbf{57.93}} & \textbf{0.683} &	55.28	& \multicolumn{1}{c|}{\textbf{54.13}} & \textbf{0.720} &	45.90 &	\textbf{49.96} \\ 
					\hline
					\multicolumn{1}{|c|}{{\small{Swad \cite{cha2021swad} + MSP}}}& 0.570 &	60.52 &	\multicolumn{1}{c|}{42.85} & 0.661 &	53.49 &	\multicolumn{1}{c|}{51.06} & 0.661 &	47.90 &	49.10 \\
					\multicolumn{1}{|c|}{{\small{Swad+\textbf{\OUR}}}}&  \textbf{0.700} &	60.52 &	\multicolumn{1}{c|}{\textbf{57.05}}  & \textbf{0.682} &	53.49 &	\multicolumn{1}{c|}{\textbf{52.92}}  &  \textbf{0.682} &	47.90 &	\textbf{50.73}\\ 
					\hline
				\end{tabular}
			}
			\label{tab:open_dg_new}
		\end{table}
	\end{minipage}%
\end{figure}
\begin{table}[t]
\centering
\caption{Results obtained by changing the configuration of the relational module. We compare \OUR with handcrafted feature aggregation strategies for sample pairs.}
\resizebox{\textwidth}{!}{
\centering

\begin{tabular}{cc|cccccccc||cc|}
\cline{3-12}
& &  \multicolumn{10}{c|}{\small{PACS - \textbf{Multi-Source}}} \\
\cline{3-12} 
& & \multicolumn{2}{c|}{{\small{ArtPainting}}}  & \multicolumn{2}{c|}{{\small{Sketch}}}  & \multicolumn{2}{c|}{{\small{Cartoon}}} & \multicolumn{2}{c||}{{\small{Photo}}} & \multicolumn{2}{c|}{{\small{\textbf{Avg.}}}} \\
 & & \scriptsize{AUROC} $\uparrow$ & \multicolumn{1}{c|}{\scriptsize{FPR95} $\downarrow$} &  \scriptsize{AUROC} $\uparrow$ & \multicolumn{1}{c|}{\scriptsize{FPR95} $\downarrow$} & \scriptsize{AUROC} $\uparrow$ & \scriptsize{FPR95} $\downarrow$  & \multicolumn{1}{|c}{\scriptsize{AUROC}} $\uparrow$ & \scriptsize{FPR95} $\downarrow$  & \multicolumn{1}{c}{\scriptsize{AUROC}} $\uparrow$ & \scriptsize{FPR95} $\downarrow$  \\

\hline

\multicolumn{2}{|c|}{\multirow{1}{*}{\small{\textbf{\OUR}}}} & 0.750 &	\multicolumn{1}{c|}{0.820} &	0.685	 & \multicolumn{1}{c|}{0.894} & 0.660 &	\multicolumn{1}{c|}{0.854} & 0.963 &	\multicolumn{1}{c||}{0.181} & \textbf{0.765} &	\textbf{0.687} \\

\hline

\multicolumn{1}{|c}{\multirow{3}{*}{Aggreg.}}&  \small{Max}  & 0.676 &	\multicolumn{1}{c|}{0.899} &	0.785 &	\multicolumn{1}{c|}{0.742} &	0.616 &	 \multicolumn{1}{c|}{0.940} & 0.827 &	\multicolumn{1}{c||}{0.786} &	0.726 &	0.842 \\

\multicolumn{1}{|c}{} &   \small{Sum}  & 0.583 &	\multicolumn{1}{c|}{0.976} &	0.446 & \multicolumn{1}{c|}{0.988} & 0.514 &	 \multicolumn{1}{c|}{0.996} & 0.575 &	\multicolumn{1}{c||}{1.000} & 0.530	& 0.990  \\
\multicolumn{1}{|c}{} &    \small{Concat}  & 0.676 &	\multicolumn{1}{c|}{0.842} &	0.710 &	\multicolumn{1}{c|}{0.790} & 0.635 &	\multicolumn{1}{c|}{0.902} & 0.921 &	\multicolumn{1}{c||}{0.438} & 0.736 &	0.743 \\

\hline

\end{tabular}

}
\vspace{-3mm}

\label{tab:ablation} 
\end{table}

The good performance obtained by \OUR in the analyzed settings suggests that it could be directly and successfully applied in various real-world tasks. 
We focus on the challenging open-set DG problem that was introduced in \cite{shu2021_OPENDG} (see Fig. \ref{fig:opendg_setting}). Multiple source domains are combined together and their different label sets cause some classes to exist in many more domains than other classes. The target is drawn from a different distribution with a large shift with respect to the source, both in terms of style and semantic content. Indeed, the target contains more classes than the source and they should be identified as unknown at test time. 
Existing closed-set DG methods are able to learn classification models that generalize to the unseen target domain containing the same categories of the source. 
One simple way to let them reject samples of novel classes is to add a threshold on MSP, considering unknown the samples with uncertain predictions, as done in DAML. 
We can apply the same technique on SagNet, Diversify and SWAD. Still, the results can take further  advantage from a method better suited to spot semantic novelties across domains, as \OUR. 

We consider the source domains as support set and the target as test, running the evaluation procedure of \OUR to obtain the normality score for each target sample. The obtained values are combined with the MSP produced by each reference method with a simple score averaging as an ensemble strategy. Since the two normality evaluations originate from different input features we aim at leveraging their complementary nature and maximize the final unknown rejection accuracy. 
The obtained results are shown in Table \ref{tab:open_dg_new}. 
In all cases, integrating \OUR with the other methods provides an improvement both in AUROC and in H-score, with Acc maintaining the exact same values, as \OUR does not influence predictions on known classes.
\section{Further analysis and discussions}

\textbf{Learnable Relational Module.} %
To assess the influence of our design choices for the relational module in \OUR, we consider alternative strategies to combine the features of sample pairs. 
Specifically, we evaluate the effect of substituting our transformer-based relational module with hand-designed aggregation functions (\emph{Max}/\emph{Sum}/\emph{Concat}), followed by an MLP whose output is fed to the final semantic similarity head. 
The MLP module is designed to have a similar number of learnable parameters with respect to our transformer-based one. 
For \emph{Concat} we exploit the feature concatenation as already done in \cite{patacchiola2020_SELFrelreason}. Note that the \emph{permutation invariance property} of our transformer 
gets lost by feature concatenation: the order of the images in the pair influences the final predictions. 

\rev{Table \ref{tab:ablation} reports the results of this analysis on the PACS multi-source setting.}
We argue that the superior performance of \OUR originates from having learned the feature aggregation function rather than relying on a fixed approach imposed a priori. Still, \textit{Max} and \textit{Concat} are able to obtain quite good results (better than what was obtained by the second best in Table \ref{tab:finetuning_new},  OODFormer \cite{koner2021oodformer} Avg$_{AUROC}$: 0.705, Avg$_{FPR95}$: 0.869). \rev{This is}
an additional evidence of the effectiveness of the relational reasoning approach for semantic novelty detection. 

We remark that an important characteristics of \OUR is its ability to learn jointly the feature embedding and the semantic similarity metric through an end-to-end training. As highlighted by Sung et al. \cite{sung2018_LearningToCompare} this is a superior strategy with respect to both methods that learn the feature embeddings but use a fixed similarity measure (e.g. Euclidean) \cite{fontanel2020boosting}, and methods that instead learn a similarity measure on top of a fixed feature representation \cite{mensink2012metric,chen2012bayesian}.
\smallskip

\noindent\textbf{Regression vs Classification.} 
\rev{As mentioned in Sec. 3.2, the relational reasoning learning paradigm can be cast as both a binary classification and a regression problem. We believe the latter is more conceptually appropriate as we want to learn a semantic similarity measure with a continuous value.}
\rev{The alternative solution consists in a binary \emph{same} vs \emph{different} task, in which the prediction for the class \emph{same} could be used as semantic similarity measure. In practice, what really differentiates the two approaches is the trend of the loss function.}

In Fig. \ref{fig:CE_vs_regression_trend} we represent the loss when varying the probability assigned to the correct class for both the classification cross entropy (CE) and the regression MSE. 
In both cases a high loss is assigned to a low probability and vice-versa. 
\rev{In the very small and rarely populated region of low probability values ($p \approx 0$), CE is higher than MSE.}
\rev{While the MSE gives more importance through higher loss values to hard samples belonging to the intermediate probability region, the CE focuses more on easy samples ($p > 0.75$) pushing their already high probability values to the same even higher output}.
The final effect of the CE is a minimization of the difference among the samples, which is not ideal when we want to use the confidence as a semantic similarity metric.

We compare the performance obtained by \OUR with the two different choices for the loss in Fig. \ref{fig:CE_vs_MSE_performance}. We considered all the dataset benchmarks already used for the open-set DG analysis and we show how both the losses provide good results, with the regression outperforming the classification one in all the cases.

\definecolor{cornflowerblue}{rgb}{0.003, 0.522, 0.75}
\definecolor{red_}{rgb}{0.89, 0.137, 0.0}

\begin{figure}[t]
	\begin{minipage}[]{0.27\textwidth}
	    \begin{figure}[H]
	    \centering
	    \resizebox{1\linewidth}{!}{
	    \begin{tikzpicture}
    \begin{axis}[grid style=dashed,xmin=-0.001,xmax=1,ymin=0,ymax=5,xtick={0,0.25,0.5,0.75,1},unit vector ratio=4 1,ymajorgrids=true,ylabel={Loss},
    y label style={at={(axis description cs:0.2,0.5)},anchor=south}, xlabel={Pred. probability $p$},legend image post style={scale=0.3}]
        \addplot[
        color=red_,
            domain = -0.5:1,
            samples = 300,
            smooth,
            thick,
        ] {-ln(x)};
            \addplot[
            color=cornflowerblue,
            domain = -0.5:1,
            samples = 300,
            smooth,
            thick,
        ] {((x-1)/0.5)^2};
    \addlegendentry{Classification Loss}
     \addlegendentry{Regression Loss}
    \end{axis}
    \end{tikzpicture}
    }
    \caption{Loss trend for the probability of the correct class.}
    \label{fig:CE_vs_regression_trend}
	\end{figure}
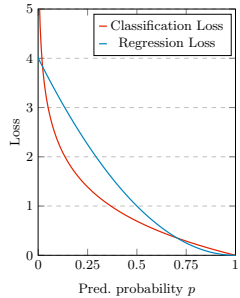
	\end{minipage}%
	\hspace{5mm}
	\begin{minipage}[]{0.66\textwidth}
    \begin{figure}[H]
	\includegraphics[width=1.0\linewidth]{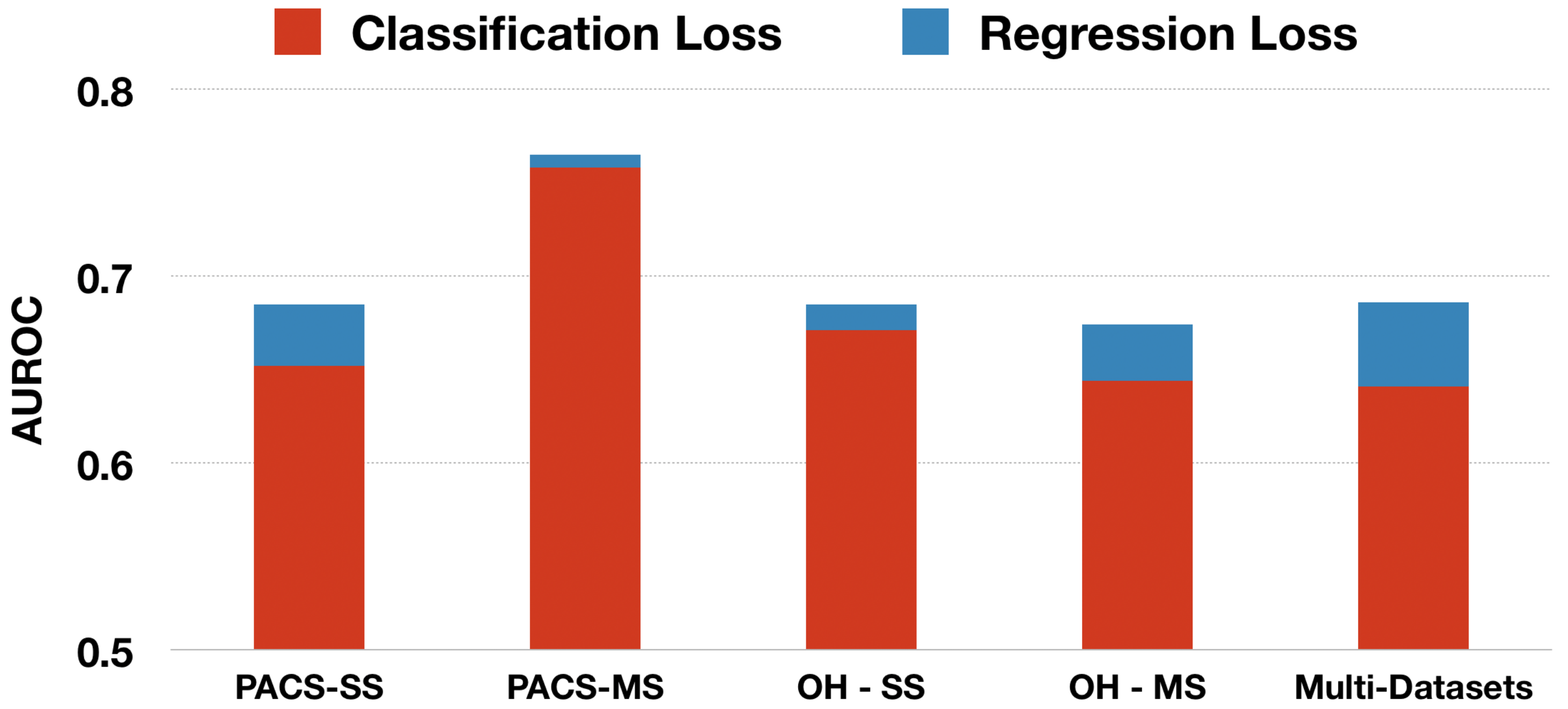}
	\caption{AUROC comparison with \OUR trained for classification via Cross Entropy Loss or for Regression via MSE. OH stands for Office-Home. SS and MS indicate respectively the Single- and Multi-Source settings.\label{fig:CE_vs_MSE_performance}}
	\end{figure}
	\end{minipage}
\end{figure}

\section{Conclusions}

\rev{In this paper we analyzed the problem of semantic novelty detection by extensively studying how traditional representation learning methods can be used for this task. Moreover, we introduced \OUR a representation learning approach that exploits relational reasoning to model semantic similarity among pairs of samples. \OUR exploits a basic transformer architecture and, once trained on ImageNet1k, it allows to identify whether a test sample belongs to a known or an unknown category by simply comparing it with the reference support set without the need for finetuning.
Our thorough experimental analysis has demonstrated the effectiveness of \OUR in both intra- and cross-domain settings, and its potential as plug-and-play module 
to transform closed-set domain generalization approaches into reliable open-set methods with state-of-the-art results.}

A trustworthy semantic novelty detection method that is able to prevent wrong annotations by identifying unknown categories without any training time latency is a crucial component in many real-world applications. We believe that our work can pave the way for more research in this direction, \rev{focusing on novel paradigms or more advanced architectures for relational reasoning.}

\smallskip\noindent\textbf{Acknowledgements} Computational resources for this work were provided by IIT (HPC infrastructure). We also
acknowledge the CINECA award IsC94 Tr-OSDG under
the ISCRA initiative, for the availability of high performance
computing resources and support. We also acknowledge the support of the European H2020 Elise project (\url{www.elise-ai.eu}).

\appendix

\begin{algorithm}[tb]
\caption{ReSeND train procedure}
\label{alg:train}
\begin{algorithmic}
\Require  $\sS$, $\sT$, $f_{\theta}:\R^{H \times W \times C} \to \R^d$, $r_{\gamma}$, $c_{\delta}$
\Procedure{create\_pairs}{$\sS$}
\State $pairs = []$
\For{\textbf{each} $(\bx^s,y^s$) \textbf{in} $\{\sS\}$}
\State $pairs.append((\textit{rand\_same\_class}(y^s), \bx^s, 1))$
\State $pairs.append((\mathit{rand\_diff\_class}(y^s), \bx^s, 0))$
\EndFor
\State \textbf{return} pairs
\EndProcedure

\Procedure{main()}{}
\For{$epoch$ \textbf{in} $ range(n\_epochs)$ }
\State $pairs = create\_pairs(\sS)$
\State $\mathit{shuffle}(pairs)$
\For{$iter$ \textbf{in} $ range(iters\_epoch)$ }

\State $pairs\_batch = next\_batch(pairs)$
\State $\bx_1, \bx_2, labels = pairs\_batch$
\State $\bz_1 = f_\theta(\bx_1)$
\State $\bz_2 = f_\theta(\bx_2)$
\State $feats\_pairs = (\bz_1, \bz_2)$
\State $predictions  = c_\delta(r_\gamma(feats\_pairs))$
\State $\textit{MSE\_loss} = \mathcal{L}(predictions,labels)$  {\color{blue} \Comment{ Eq. \ref{eq:loss}}} 
\State \textbf{Update} $\theta, \gamma, \delta \gets \nabla \textit{MSE\_loss}$ 

\EndFor
\EndFor
\EndProcedure
\end{algorithmic}
\end{algorithm}
\begin{algorithm}[tb]
\caption{ReSeND eval procedure}
\label{alg:eval}
\begin{algorithmic}
\Require  $\sS$, $\sT$, $f_{\theta}:\R^{H \times W \times C} \to \R^d$, $r_{\gamma}$, $c_{\delta}$
\Procedure{compute\_prototypes}{$\sS$}
\State $prototypes = zeros((|\sY_s|,))$
\State $counters = zeros(|\sY_s|)$
\For{\textbf{each} $(\bx^s,y^s$) \textbf{in} $\{\sS\}$}
\State $\bz^s = f_\theta(\bx^s)$
\State $prototypes[y^s] += \bz^s$
\State $counters[y^s] += 1$
\EndFor
\For{$i$ \textbf{in} $range(|\sY_s|)$}
\State $prototypes[i] /= counters[i]$
\EndFor
\State \textbf{return} $prototypes$
\EndProcedure

\Procedure{main()}{}
\State $normality\_scores = [] $
\State $prototypes = compute\_prototypes(\sS)$
\For{\textbf{each} $\bx^t$ \textbf{in} $\{\sT\}$}
\State $\bz^t = f_\theta(\bx^t)$
\State $pairs = (prototypes, \bz^t.repeat())$
\State $predictions = c_\delta(r_\gamma(pairs))$
\State $score = max(softmax(predictions))$
\State $normality\_scores.append(score)$

\EndFor
\EndProcedure

\end{algorithmic}
\end{algorithm}

\section{Implementation details}
\label{ref:app_imp_details}
We start from a standard ResNet-18 \cite{he2016deep}, pretrained on ImageNet1k \cite{deng2009imagenet}, which we use as feature extractor $f_\theta$ by removing the original final classification layer. 
Our relational module $r_\gamma$ has the same structure of the transformer in ViT \cite{ViT_dosovitskiy2021an}: we use $4$ multi-head self-attention encoder blocks, a number that allows to trade-off performance and time complexity (the number of blocks highly influences the total number of learnable parameters of the network).
The features extracted by the backbone are passed through an FC projection layer before entering the transformer. \rev{The transformer input sequence is obtained concatenating the learnable label token and the representations of a pair of samples $[\bz_{l},\bz_i,\bz_j]$. The output token $\bv_l$ is then selected and passed through a final FC layer which represents the regression head $c_\delta$}. 

\noindent \rev{The transformer procedure is summarized in the following equations:\vspace{-1mm}
\begin{align}
    \bz^0 &= [ \bz_\text{l}; \, \bz_i; \, \bz_j] \label{eq:embedding} \\
    \tilde{\bz}^b &= \text{MSA}(\text{LN}(\bz^{b-1})) + \bz^{b-1}, && b=1\ldots B \label{eq:msa_apply} \\
    \bz^b &= \text{MLP}(\text{LN}(\tilde{\bz}^{b})) + \tilde{\bz}^{b}, && b=1\ldots B  \label{eq:mlp_apply} \\
    \bv_{l} &= \text{LN}(\bz^B_{l}) \label{eq:final_rep}~.\vspace{-3mm}
\end{align}}
\noindent We train our network on ImageNet1k in an end-to-end manner using the MSE loss (Eq. \ref{eq:loss} in the main paper) applied to the output of the regression head. Our training procedure uses $13$k iterations with a batch size of $4096$, where each element of the batch is an image pair. The learning rate uses a linear warmup for 500 iterations and then is fixed to $0.008$. We use LARS optimizer \cite{LARS} with momentum $0.9$ and weight decay $5\cdot10^{-5}$.
We build image pairs by selecting each image of the dataset as anchor and associating it with a randomly chosen sample with the same label to create \textit{positive} pairs and samples of different labels to create \textit{negative} pairs. All experimental results are averaged over three runs.\\
\noindent We summarize in Algorithm \ref{alg:train} and \ref{alg:eval} the training and evaluation procedure of \OUR.

\definecolor{cornflowerblue}{rgb}{0.39, 0.58, 0.93}

\begin{figure}[tb]
\centering
\begin{minipage}[t]{0.7\textwidth}

     \resizebox {\columnwidth} {!} {
        \begin{tikzpicture}
        
        \tikzstyle{every node}=[font=\scriptsize]
        \pgfplotsset{scaled x ticks=false,every axis legend/.append style={
        at={(0.5,1.03)},
        anchor=south}}

        \begin{axis}[
          enlargelimits=false,
          ylabel={AUROC},
          x label style={at={(axis description cs:0.5,0.2)},anchor=north},
          y label style={at={(axis description cs:0.13,0.5)},anchor=south},
          xlabel={ $ \times 10^6 \; image \; pairs  $},
           xmin=0, xmax=53,
           ymin=0.5, ymax=0.8,
          xtick={0,10,20,30,40,50},
          ytick={0.5, 0.6,0.7, 0.8},
          ymajorgrids=true,
          grid style=dashed,
          width=7cm,
          height=3cm,
          legend columns=-1,
        legend style={at={(0.7,0.1)},draw=none},
        every axis plot/.append style={ultra thick},
        every mark/.append style={mark size=50pt},
        label style={font=\scriptsize},
        legend style={font=\scriptsize},
        legend image post style={scale=0.3}
        ]

         \addplot[
          color=cornflowerblue,
          mark=none]
        table[x index=0,y index=1,col sep=comma]
        {numbers.txt};
        
        \addlegendentry{PACS - Multi-Source}
        \end{axis}
        \end{tikzpicture}%
      }

\end{minipage}

\caption{\label{pairs} Performance trend increasing the number of image pairs. } 
\end{figure}
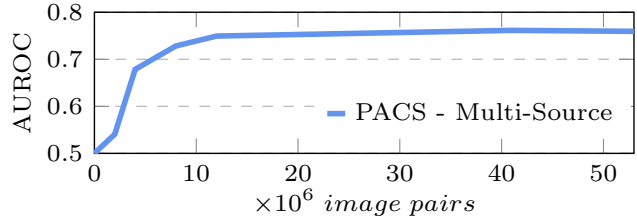

\section{Further Analysis}
\label{ref:app_results}

\textbf{Number of image pairs}. Our learning objective is based on the use of image pairs randomly created at training time by coupling samples from the training dataset. Even if the total number of image pairs that could be formed from ImageNet1k dataset is very high ($\sim 820\times10^9$), in Fig. \ref{pairs} we show that ReSeND converges after having seen a relatively small portion of them.

\bibliographystyle{splncs04}
\bibliography{egbib}

\begin{thebibliography}{10}
\providecommand{\url}[1]{\texttt{#1}}
\providecommand{\urlprefix}{URL }
\providecommand{\doi}[1]{https://doi.org/#1}

\bibitem{relationalbias_arxiv}
Battaglia, P., Hamrick, J.B.C., Bapst, V., Sanchez, A., Zambaldi, V.,
  Malinowski, M., Tacchetti, A., Raposo, D., Santoro, A., Faulkner, R.,
  Gulcehre, C., Song, F., Ballard, A., Gilmer, J., Dahl, G.E., Vaswani, A.,
  Allen, K., Nash, C., Langston, V.J., Dyer, C., Heess, N., Wierstra, D.,
  Kohli, P., Botvinick, M., Vinyals, O., Li, Y., Pascanu, R.: Relational
  inductive biases, deep learning, and graph networks. arXiv:1806.01261  (2018)

\bibitem{BengioTPAMI2013}
Bengio, Y., Courville, A., Vincent, P.: Representation learning: A review and
  new perspectives. IEEE TPAMI  \textbf{35}(8),  1798–1828 (aug 2013)

\bibitem{Bergman2020Classification-Based}
Bergman, L., Hoshen, Y.: Classification-based anomaly detection for general
  data. In: ICLR (2020)

\bibitem{bird2009natural}
Bird, S., Klein, E., Loper, E.: Natural language processing with Python:
  analyzing text with the natural language toolkit. O'Reilly Media, Inc. (2009)

\bibitem{bucci2020ros}
Bucci, S., Loghmani, M.R., Tommasi, T.: On the effectiveness of image rotation
  for open set domain adaptation. In: ECCV (2020)

\bibitem{caron2020unsupervised}
Caron, M., Misra, I., Mairal, J., Goyal, P., Bojanowski, P., Joulin, A.:
  Unsupervised learning of visual features by contrasting cluster assignments.
  In: NeurIPS (2020)

\bibitem{cha2021swad}
Cha, J., Chun, S., Lee, K., Cho, H.C., Park, S., Lee, Y., Park, S.: Swad:
  Domain generalization by seeking flat minima. In: NeurIPS (2021)

\bibitem{chen2012bayesian}
Chen, D., Cao, X., Wang, L., Wen, F., Sun, J.: Bayesian face revisited: A joint
  formulation. In: ECCV (2012)

\bibitem{chen2021atom}
Chen, J., Li, Y., Wu, X., Liang, Y., Jha, S.: Atom: Robustifying
  out-of-distribution detection using outlier mining. In: ECML (2021)

\bibitem{simclr2020}
Chen, T., Kornblith, S., Norouzi, M., Hinton, G.: A simple framework for
  contrastive learning of visual representations. In: ICML (2020)

\bibitem{Chen_2021_CVPRsimsiam}
Chen, X., He, K.: Exploring simple siamese representation learning. In: CVPR
  (2021)

\bibitem{Cheng2020_AudioVisualTransf}
Cheng, Y., Wang, R., Pan, Z., Feng, R., Zhang, Y.: Look, listen, and attend:
  Co-attention network for self-supervised audio-visual representation
  learning. In: ACM Multimedia (2020)

\bibitem{Cimpoi_2014_CVPR_textures}
Cimpoi, M., Maji, S., Kokkinos, I., Mohamed, S., Vedaldi, A.: Describing
  textures in the wild. In: CVPR (2014)

\bibitem{coates2011analysis}
Coates, A., Ng, A., Lee, H.: An analysis of single-layer networks in
  unsupervised feature learning. In: AISTATS (2011)

\bibitem{AE_anomaly_ICPR20}
Collin, A.S., De~Vleeschouwer, C.: Improved anomaly detection by training an
  autoencoder with skip connections on images corrupted with stain-shaped
  noise. In: ICPR (2021)

\bibitem{deeckeICML21}
Deecke, L., Ruff, L., Vandermeulen, R.A., Bilen, H.: Transfer-based semantic
  anomaly detection. In: ICML (2021)

\bibitem{deng2009imagenet}
Deng, J., Dong, W., Socher, R., Li, L.J., Li, K., Fei-Fei, L.: Imagenet: A
  large-scale hierarchical image database. In: CVPR (2009)

\bibitem{ViT_dosovitskiy2021an}
Dosovitskiy, A., Beyer, L., Kolesnikov, A., Weissenborn, D., Zhai, X.,
  Unterthiner, T., Dehghani, M., Minderer, M., Heigold, G., Gelly, S.,
  Uszkoreit, J., Houlsby, N.: An image is worth 16x16 words: Transformers for
  image recognition at scale. In: ICLR (2021)

\bibitem{du2021curious}
Du, Y., Gan, C., Isola, P.: Curious representation learning for embodied
  intelligence. In: ICCV (2021)

\bibitem{Ericsson2021HowTransfer}
Ericsson, L., Gouk, H., Hospedales, T.M.: {How Well Do Self-Supervised Models
  Transfer?} In: CVPR (2021)

\bibitem{fontanel2020boosting}
Fontanel, D., Cermelli, F., Mancini, M., Bulo, S.R., Ricci, E., Caputo, B.:
  Boosting deep open world recognition by clustering. IEEE RAL  \textbf{5}(4),
  5985--5992 (2020)

\bibitem{GenOpenMax}
Ge, Z., Demyanov, S., Chen, Z., Garnavi, R.: Generative openmax for multi-class
  open set classification. In: BMVC (2017)

\bibitem{gidaris2018unsupervised}
Gidaris, S., Singh, P., Komodakis, N.: Unsupervised representation learning by
  predicting image rotations. In: ICLR (2018)

\bibitem{golan2018deep}
Golan, I., El-Yaniv, R.: Deep anomaly detection using geometric
  transformations. In: NeurIPS (2018)

\bibitem{NIPS2014gan}
Goodfellow, I., Pouget-Abadie, J., Mirza, M., Xu, B., Warde-Farley, D., Ozair,
  S., Courville, A., Bengio, Y.: Generative adversarial nets. In: NeurIPS
  (2014)

\bibitem{GoodBengBook}
Goodfellow, I.J., Bengio, Y., Courville, A.: Deep Learning. MIT Press (2016)

\bibitem{MOCO_He_2020_CVPR}
He, K., Fan, H., Wu, Y., Xie, S., Girshick, R.: Momentum contrast for
  unsupervised visual representation learning. In: CVPR (2020)

\bibitem{he2016deep}
He, K., Zhang, X., Ren, S., Sun, J.: Deep residual learning for image
  recognition. In: CVPR (2016)

\bibitem{hendrycks17baseline}
Hendrycks, D., Gimpel, K.: A baseline for detecting misclassified and
  out-of-distribution examples in neural networks. In: ICLR (2017)

\bibitem{hendrycks2018deep}
Hendrycks, D., Mazeika, M., Dietterich, T.: Deep anomaly detection with outlier
  exposure. In: ICLR (2019)

\bibitem{HintonSalakhutdinov2006b}
Hinton, G.E., Salakhutdinov, R.R.: Reducing the dimensionality of data with
  neural networks. Science  \textbf{313}(5786),  504--507 (Jul 2006)

\bibitem{hjelm2018learning}
Hjelm, R.D., Fedorov, A., Lavoie-Marchildon, S., Grewal, K., Bachman, P.,
  Trischler, A., Bengio, Y.: Learning deep representations by mutual
  information estimation and maximization. In: ICLR (2019)

\bibitem{gODIN2020}
{Hsu}, Y.C., {Shen}, Y., {Jin}, H., {Kira}, Z.: Generalized odin: Detecting
  out-of-distribution image without learning from out-of-distribution data. In:
  CVPR (2020)

\bibitem{Hu_2018_CVPR}
Hu, H., Gu, J., Zhang, Z., Dai, J., Wei, Y.: Relation networks for object
  detection. In: CVPR (2018)

\bibitem{huang2021importance}
Huang, R., Geng, A., Li, Y.: On the importance of gradients for detecting
  distributional shifts in the wild. NeurIPS  (2021)

\bibitem{huang2021_MOS}
Huang, R., Li, Y.: Mos: Towards scaling out-of-distribution detection for large
  semantic space. In: CVPR (2021)

\bibitem{Jenni_steering}
Jenni, S., Jin, H., Favaro, P.: Steering self-supervised feature learning
  beyond local pixel statistics. In: CVPR (2020)

\bibitem{Johnson_clevr}
Johnson, J., Hariharan, B., van~der Maaten, L., Fei{-}Fei, L., Zitnick, C.L.,
  Girshick, R.B.: {CLEVR:} {A} diagnostic dataset for compositional language
  and elementary visual reasoning. In: CVPR (2017)

\bibitem{khosla2020supervised}
Khosla, P., Teterwak, P., Wang, C., Sarna, A., Tian, Y., Isola, P., Maschinot,
  A., Liu, C., Krishnan, D.: Supervised contrastive learning. In: NeurIPS
  (2020)

\bibitem{Kim2020RaPP}
Kim, K.H., Shim, S., Lim, Y., Jeon, J., Choi, J., Kim, B., Yoon, A.S.: Rapp:
  Novelty detection with reconstruction along projection pathway. In: ICLR
  (2020)

\bibitem{Kingma2014}
Kingma, D.P., Welling, M.: {Auto-Encoding Variational Bayes}. In: ICLR (2014)

\bibitem{kolesnikov2019revisiting}
Kolesnikov, A., Zhai, X., Beyer, L.: Revisiting self-supervised visual
  representation learning. In: CVPR (2019)

\bibitem{koner2021oodformer}
Koner, R., Sinhamahapatra, P., Roscher, K., G{\"u}nnemann, S., Tresp, V.:
  Oodformer: Out-of-distribution detection transformer. In: BMVC (2021)

\bibitem{lee2018training}
{Lee}, K., {Lee}, H., {Lee}, K., {Shin}, J.: Training confidence-calibrated
  classifiers for detecting out-of-distribution samples. In: ICLR (2018)

\bibitem{NEURIPS2018_abdeb6f5}
Lee, K., Lee, K., Lee, H., Shin, J.: A simple unified framework for detecting
  out-of-distribution samples and adversarial attacks. In: NeurIPS (2018)

\bibitem{cutpaste}
Li, C.L., Sohn, K., Yoon, J., Pfister, T.: Cutpaste: Self-supervised learning
  for anomaly detection and localization. In: CVPR (2021)

\bibitem{li_2017_pacs}
Li, D., Yang, Y., Song, Y.Z., Hospedales, T.M.: Deeper, broader and artier
  domain generalization. In: ICCV (2017)

\bibitem{liang2017enhancing}
Liang, S., Li, Y., Srikant, R.: Enhancing the reliability of
  out-of-distribution image detection in neural networks. In: ICLR (2018)

\bibitem{liu2020energy}
Liu, W., Wang, X., Owens, J., Li, Y.: Energy-based out-of-distribution
  detection. NeurIPS  (2020)

\bibitem{mensink2012metric}
Mensink, T., Verbeek, J., Perronnin, F., Csurka, G.: Metric learning for large
  scale image classification: Generalizing to new classes at near-zero cost.
  In: ECCV (2012)

\bibitem{nalisnick2018do}
Nalisnick, E., Matsukawa, A., Teh, Y.W., Gorur, D., Lakshminarayanan, B.: Do
  deep generative models know what they don't know? In: ICLR (2019)

\bibitem{nam2021reducing}
Nam, H., Lee, H., Park, J., Yoon, W., Yoo, D.: Reducing domain gap by reducing
  style bias. In: CVPR (2021)

\bibitem{Neal_2018_ECCV}
Neal, L., Olson, M., Fern, X., Wong, W.K., Li, F.: Open set learning with
  counterfactual images. In: ECCV (2018)

\bibitem{Newell_2020_CVPR}
Newell, A., Deng, J.: How useful is self-supervised pretraining for visual
  tasks? In: CVPR (2020)

\bibitem{noroozi2016}
Noroozi, M., Favaro, P.: Unsupervised learning of visual representations by
  solving jigsaw puzzles. In: ECCV (2016)

\bibitem{oza2020multiple}
Oza, P., Nguyen, H.V., Patel, V.M.: Multiple class novelty detection under data
  distribution shift. In: ECCV (2020)

\bibitem{pan2021actor}
Pan, J., Chen, S., Shou, M.Z., Liu, Y., Shao, J., Li, H.: Actor-context-actor
  relation network for spatio-temporal action localization. In: CVPR (2021)

\bibitem{PAPADOPOULOS2021138}
Papadopoulos, A.A., Rajati, M.R., Shaikh, N., Wang, J.: Outlier exposure with
  confidence control for out-of-distribution detection. Neurocomputing
  \textbf{441},  138--150 (2021)

\bibitem{park2020learning}
Park, H., Noh, J., Ham, B.: Learning memory-guided normality for anomaly
  detection. In: CVPR (2020)

\bibitem{patacchiola2020_SELFrelreason}
Patacchiola, M., Storkey, A.: Self-supervised relational reasoning for
  representation learning. In: NeurIPS (2020)

\bibitem{peng2019moment_domainnet}
Peng, X., Bai, Q., Xia, X., Huang, Z., Saenko, K., Wang, B.: Moment matching
  for multi-source domain adaptation. In: ICCV (2019)

\bibitem{peng2017visda}
Peng, X., Usman, B., Kaushik, N., Hoffman, J., Wang, D., Saenko, K.: Visda: The
  visual domain adaptation challenge. arXiv preprint arXiv:1710.06924  (2017)

\bibitem{Raposo_iclr17}
Raposo, D., Santoro, A., Barrett, D.G.T., Pascanu, R., Lillicrap, T.,
  Battaglia, P.W.: Discovering objects and their relations from entangled scene
  representations. In: ICLR Workshop (2017)

\bibitem{anomaly_detection_survey_dietterich}
Ruff, L., Kauffmann, J.R., Vandermeulen, R.A., Montavon, G., Samek, W., Kloft,
  M., Dietterich, T.G., Müller, K.R.: A unifying review of deep and shallow
  anomaly detection. Proceedings of the IEEE  \textbf{109}(5),  756--795 (2021)

\bibitem{saenko2010adapting}
Saenko, K., Kulis, B., Fritz, M., Darrell, T.: Adapting visual category models
  to new domains. In: ECCV (2010)

\bibitem{Santoro_nips18}
Santoro, A., Faulkner, R., Raposo, D., Rae, J.W., Chrzanowski, M., Weber, T.,
  Wierstra, D., Vinyals, O., Pascanu, R., Lillicrap, T.P.: Relational recurrent
  neural networks. In: NeurIPS (2018)

\bibitem{Santoro_nips17}
Santoro, A., Raposo, D., Barrett, D.G.T., Malinowski, M., Pascanu, R.,
  Battaglia, P.W., Lillicrap, T.: A simple neural network module for relational
  reasoning. In: NeurIPS (2017)

\bibitem{gram_ood}
Sastry, C.S., Oore, S.: Detecting out-of-distribution examples with {G}ram
  matrices. In: ICML (2020)

\bibitem{Segaran07}
Segaran, T.: Programming Collective Intelligence: Building Smart Web 2.0
  Applications. O'Reilly (2007)

\bibitem{sehwag2021_SSD}
Sehwag, V., Chiang, M., Mittal, P.: Ssd: A unified framework for
  self-supervised outlier detection. In: ICLR (2021)

\bibitem{aaai_SensoyKCS20}
Sensoy, M., Kaplan, L.M., Cerutti, F., Saleki, M.: Uncertainty-aware deep
  classifiers using generative models. In: AAAI (2020)

\bibitem{shu2021_OPENDG}
Shu, Y., Cao, Z., Wang, C., Wang, J., Long, M.: Open domain generalization with
  domain-augmented meta-learning. In: CVPR (2021)

\bibitem{fort2021_ExploringLimitsOOD}
Stanislav~Fort, J.R., Lakshminarayanan, B.: Exploring the limits of
  out-of-distribution detection. In: NeurIPS (2021)

\bibitem{sung2018_LearningToCompare}
Sung, F., Yang, Y., Zhang, L., Xiang, T., Torr, P.H., Hospedales, T.M.:
  Learning to compare: Relation network for few-shot learning. In: CVPR (2018)

\bibitem{tack2020_CSI}
Tack, J., Mo, S., Jeong, J., Shin, J.: Csi: Novelty detection via contrastive
  learning on distributionally shifted instances. In: NeurIPS (2020)

\bibitem{DeiT_pmlr-v139-touvron21a}
Touvron, H., Cord, M., Douze, M., Massa, F., Sablayrolles, A., Jegou, H.:
  Training data-efficient image transformers \& distillation through attention.
  In: ICML (2021)

\bibitem{venkateswara2017deep}
Venkateswara, H., Eusebio, J., Chakraborty, S., Panchanathan, S.: Deep hashing
  network for unsupervised domain adaptation. In: CVPR (2017)

\bibitem{wang2021learning}
Wang, Z., Luo, Y., Qiu, R., Huang, Z., Baktashmotlagh, M.: Learning to
  diversify for single domain generalization. In: ICCV (2021)

\bibitem{winkens2020contrastive}
Winkens, J., Bunel, R., Roy, A.G., Stanforth, R., Natarajan, V., Ledsam, J.R.,
  MacWilliams, P., Kohli, P., Karthikesalingam, A., Kohl, S., Cemgil, T.,
  Eslami, S.M.A., Ronneberger, O.: Contrastive training for improved
  out-of-distribution detection. arXiv:2007.05566  (2020)

\bibitem{xia2020synthesize}
Xia, Y., Zhang, Y., Liu, F., Shen, W., Yuille, A.: Synthesize then compare:
  Detecting failures and anomalies for semantic segmentation. In: ECCV (2020)

\bibitem{Yang_2021_ICCV}
Yang, J., Wang, H., Feng, L., Yan, X., Zheng, H., Zhang, W., Liu, Z.:
  Semantically coherent out-of-distribution detection. In: ICCV (2021)

\bibitem{LARS}
You, Y., Gitman, I., Ginsburg, B.: Large batch training of convolutional
  networks. arXiv:1708.03888  (2017)

\bibitem{yun2019cutmix}
Yun, S., Han, D., Oh, S.J., Chun, S., Choe, J., Yoo, Y.: Cutmix: Regularization
  strategy to train strong classifiers with localizable features. In: ICCV
  (2019)

\bibitem{zambaldi2018deep}
Zambaldi, V., Raposo, D., Santoro, A., Bapst, V., Li, Y., Babuschkin, I.,
  Tuyls, K., Reichert, D., Lillicrap, T., Lockhart, E., Shanahan, M., Langston,
  V., Pascanu, R., Botvinick, M., Vinyals, O., Battaglia, P.: Deep
  reinforcement learning with relational inductive biases. In: ICLR (2019)

\bibitem{barlow_ZbontarJMLD21}
Zbontar, J., Jing, L., Misra, I., LeCun, Y., Deny, S.: Barlow twins:
  Self-supervised learning via redundancy reduction. In: ICML (2021)

\bibitem{Zhang_2021_CVPR}
Zhang, H., Koniusz, P., Jian, S., Li, H., Torr, P.H.S.: Rethinking class
  relations: Absolute-relative supervised and unsupervised few-shot learning.
  In: CVPR (2021)

\end{thebibliography}
\end{document}